\pdfoutput=1
\documentclass[11pt]{article}

\usepackage[final]{EMNLP2023}

\usepackage{times}
\usepackage{latexsym}
\usepackage{verbatim}
\usepackage{amsmath}
\usepackage{amssymb}

\usepackage[T1]{fontenc}
\usepackage[utf8]{inputenc}

\usepackage{microtype}

\usepackage{inconsolata}

\usepackage{graphicx}
\usepackage{tablefootnote}

\usepackage{siunitx}
\usepackage{booktabs}
\usepackage{hyperref}
\usepackage{float}
\usepackage{multirow}
\usepackage[multiple]{footmisc}

\title{Leveraging Large Language Models for Code-Mixed Data Augmentation in Sentiment Analysis}

\author{Linda Zeng \\
The Harker School \\
  500 Saratoga Ave, San Jose, CA 95129 \\
  \texttt{26lindaz@students.harker.org}}

\begin{document}
\maketitle
\begin{abstract}

Code-mixing (CM), where speakers blend languages within a single expression, is prevalent in multilingual societies but poses challenges for natural language processing due to its complexity and limited data. We propose using a large language model to generate synthetic CM data, which is then used to enhance the performance of task-specific models for CM sentiment analysis. Our results show that in Spanish-English, synthetic data improved the F1 score by 9.32\%, outperforming previous augmentation techniques. However, in Malayalam-English, synthetic data only helped when the baseline was low; with strong natural data, additional synthetic data offered little benefit. Human evaluation confirmed that this approach is a simple, cost-effective way to generate natural-sounding CM sentences, particularly beneficial for low baselines. Our findings suggest that few-shot prompting of large language models is a promising method for CM data augmentation and has significant impact on improving sentiment analysis, an important element in the development of social influence systems.

\begin{comment}
Code-mixing (CM), where a speaker blends two or more languages within a single expression, is integral in multilingual societies. However, natural language processing systems struggle to handle CM due to its complexity and limited data availability. We propose employing a large language model to augment existing data, allowing downstream models to better learn CM features. We prompted GPT-4 to generate synthetic CM data, which were then used alongside scarce human data to train a task-specific model for CM sentiment analysis. We found that LLM-generated data was effective in improving performance to a certain threshold. In Spanish-English, synthetic data raised the downstream model’s F1 score by 9.32\%, outperforming past data augmentation techniques. However, in Malayalam-English, adding synthetic data only improved results when the baseline performance was low. Because our Malayalam-English baseline already surpassed the highest published benchmark by 6\% with solely natural data, it did not benefit from additional synthetic data. Through human evaluation and empirical data observations, we found that this method is a simple, cost-effective way of generating natural-sounding CM sentences, which increase performance for otherwise low baselines. Our findings offer a promising method for CM data augmentation, encouraging paths for further exploration with few-shot prompting large language models in the domain of CM. 
\end{comment}

\end{abstract}

\section{Introduction} \label{intro}

%Introducing CM and Importance

\begin{figure}[t]
  \includegraphics[width=\columnwidth]{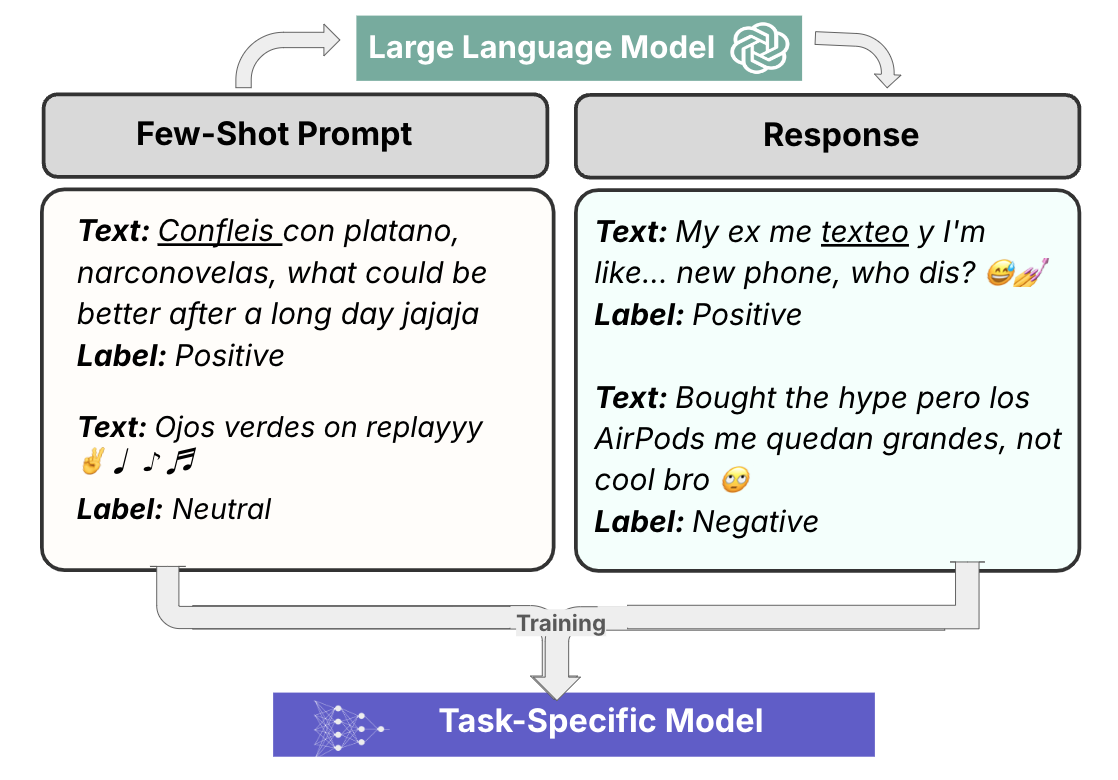}
  \caption{Overall system workflow with examples of Spanish-English CM tweets as natural data (left) and synthetic data (right). Underlined words represent Spanish-English hybrid words, examples of the complexities introduced by CM. Translations of CM sentences into English are provided in Appendix \ref{sec:translations}.}
  \label{fig:examples}
\end{figure}

%, which often prove a challenge to multilingual models. CM examples also reveal real world applications, as people reflect on songs "Ojos verdes" and products "AirPods." The parallel nature of natural and synthetic data is also demonstrated with the replay example.

%TODO: diff LLM, diff base model, implement other data aug methods to compare??, human evaluation!!, try to investigate/improve malayalam (diff prompt/diff llm/etc.) + CITE ALL ALL ALL PREVIOUS RELATED WORK (COMPARE TO OTHER WORK!! / IMPLEMENT OTHER WORK), complicated/tecnical soundness. use existing work as baseline and then use prompt-based llm to compare. clarify that the llm is used for data augmentation, use for when u have little training data and use it to augment training data!!
%- find some novel innovation

%- what is considered ablation vs normal
%- have chatgpt and professor criticize my paper!!!! 
% REAFFIRM SHOT VALUES WITH DIFF LANGUAGE?

Code-mixing (CM), or code-switching, is the practice of switching between languages within a conversation or utterance. This practice is integral to multilingual societies, particularly in Mexico and urban India \citep{parshad2016hinglish}, and is also significant in computer-mediated communication and social media, where multilingual users are predominant \citep{rijhwani-etal-2017-estimating}. Despite its ubiquity, CM is mostly spoken and found in personal messages, making training data scarce and leading to poorer Natural Language Processing (NLP) model performance compared to monolingual text \citep{pratapa-etal-2018-word, yong-etal-2023-prompting}. 

Social influence (SI) refers to the changes in thoughts, feelings, attitudes, or behaviors resulting from interactions with others. In multilingual societies, CM reflects an important aspect of these interactions, reflecting social dynamics and identity. Sentiment analysis (SA) is crucial for understanding these dynamics, as it captures the emotional nuances embedded in multilingual interactions. Furthermore, SA has become a primary CM task due to its need for complex semantic understanding and its implications for social media \citep{10.1016/j.procs.2019.11.174}, where CM is commonly present \citep{shrinivasan}. By accurately analyzing sentiment in code-mixed text, SI systems enhance their ability to interpret user intent and emotional states, enabling more meaningful interactions addressing the more diverse environments in which SI occurs. Since multilingual speakers bridge information on social media \citep{li2022language}, machines must also accurately analyze CM text to capture public opinion and disseminate news. However, current approaches fall short in handling code-mixed settings \citep{dogruoz-etal-2021-survey,aguilar-etal-2020-lince} due to data scarcity.

%To address data scarcity in CM sentiment analysis, researchers have explored generating synthetic CM data from monolingual data.

Beyond the CM domain, few-shot learning has shown promise in overcoming data scarcity, as Large Language Models (LLMs) trained on diverse tasks generalize to new ones with minimal training \citep{brown2020language, lin-etal-2022-shot, winata-etal-2021-language}. LLMs are used for data augmentation \citep{ding2024data, whitehouse-etal-2023-llm, yoo2021gpt3mix, dai2023auggpt}, training data generation \citep{yu2023large}, and knowledge distillation \citep{xu2024survey, phuong2021understanding}, particularly in low-resource settings \citep{ding2024data}. However, this approach remains underexplored in the CM domain, which presents unique challenges \citep{zhang-etal-2023-multilingual}.

In this work, we bring LLM-powered data augmentation to the task of code-mixed sentiment analysis. We use few-shot prompting to generate labeled CM SA data in Spanish-English and low-resource Malayalam-English. 
%Although we are testing this approach, it is conceivable that randomly translated data might not perform as well as LLM-generated data. This is because LLMs, which are trained on a diverse range of human-like text, can produce sentences that better reflect natural language patterns and nuances. In contrast, randomly translated data may lack this level of contextual relevance and coherence, potentially leading to less effective sentiment analysis results. 
Following \citet{li-murray-2023-zero, whitehouse-etal-2023-llm,tareq}, we quantify 
 the performance gains by fine-tuning multilingual pre-trained language models (PLMs) on the LLM-generated data. We investigate if these synthetic data samples can reflect natural code-mixing patterns and nuances compared to other data augmentation techniques and verify the synthetic data quality through human evaluation.

Figure \ref{fig:examples} displays our overall system workflow with examples of natural and synthetic data. We summarize our contributions as follows:
\begin{itemize}
    \item We introduce LLMs for CM data augmentation as a simple, cost-effective way to improve sentiment analysis models with natural-sounding sentences;
    \item We surpass past baselines, achieving third on the LinCE benchmark \citep{aguilar-etal-2020-lince} in Spanish-English and outperforming the highest published benchmark by 4.85\% on the low-resource MalayalamMixSentiment dataset \citep{chakravarthi-etal-2020-sentiment};
    \item We thoroughly analyze the efficacy of our data augmentation approach in comparison to other techniques and with human evaluation;
    \item We release the synthetic data and code on Github\footnote{\url{https://github.com/lindazeng979/LLM-CMSA}} for public use and reproducibility.
\end{itemize}

% importance of sentiment analysis
%In the work, we propose a new method to CM data augmentation in Sentiment Analysis (SA), a field important for monitoring and reflecting public opinion in social media, which is contingent on its large body of multilingual users. We harness the zero-shot and few-shot learning capabilities of Large Language Models (Brown et al., 2020) to generalize to the CM SA generation task. JUSTIFICATION SENTENCE I prompted GPT-4 to generate synthetic CM data, which were then used alongside scarce human data to train a downstream cross-lingual model for CM sentiment analysis, shown in Figure 1. This work aims to 1) introduce a new, effective LLM-powered data augmentation pipeline to improve measures of consumer sentiment in CM social media comments and 2) explore LLM's ability to generalize to the CM generation task.

%In our proposed system, we use in-context learning to prompt-engineer GPT-4 for CM Sentiment Analysis Data Augmentation. To quantify the improvement of the generated data, we combine large synthetic data and scarce natural data to finetune a pre-trained language model with a classification layer.

%We introduce a novel, efficient, bla method to mass produce synthetic CM sentences and quantify their effectiveness. My contributions include:

%MENTION ABLATION?????

%Introducing our work
% my purpose + my contribution
% 1) LLMs for two languages
% 2) new baseline for Malayalam-English
% 3) other important values

\section{Related Work}

\subsection{Data Augmentation for Code-Mixing}
Existing attempts at generating synthetic CM data focus on using linguistics theory or converting monolingual data to CM data.
 
For instance, \citet{pratapa-etal-2018-word} use Equivalence Constraint Theory to align the parse trees of Hindi and English and replace words in one language with their corresponding words in the second language. \citet{lee19d_interspeech} apply Matrix Language Frame theory to convert parallel data to CM data, and \citet{gregorius-okadome-2022-generating} use a dependency tree which predicts code-switching points and a machine translator to convert monolingual sentences to CM. While these methods consider the intention behind code-switching points \citep{solorio-liu-2008-learning}, they require expert linguistic knowledge, assume languages pairs can be parsed by the same parse tree, and rely on the accuracy of the parsers employed. % said by li and murray and by winata et al

Other approaches convert monolingual data into CM using machine translation systems \citep{vu2012speech, li2022language, tarunesh-etal-2021-machine}, word dictionaries \citep{tareq}, or parallel corpora \citep{winata-etal-2019-code, whitehouse-etal-2022-entitycs}. For instance, \citet{winata-etal-2019-code} employ a sequence-to-sequence model to learn language-switching points while \citet{chang2019codeswitching} use generative-adversarial networks. \citet{li2022language} introduce language-agnostic masks in a monolingual SA corpus to train models on recognizing the patterns of CM, and \citet{tareq} utilize word dictionaries to map monolingual data into CM SA data. Although some of these techniques account for code-switching points, they do not consistently produce natural sentences. Moreover, their effectiveness relies on the quality of the underlying systems and the assumption that large datasets with distributions similar to real CM data are available.

Unlike conversion-based methods, our approach generates CM sentences from scratch. By leveraging LLMs' multilingual pre-training and generalization capabilities, we aim to produce synthetic data that more accurately reflects the natural patterns and nuances of human-generated CM language.

\subsection{Large Language Models for Code-Mixing}

To our knowledge, LLMs have not yet been used for CM data augmentation. The closest related works are by \citet{yong-etal-2023-prompting}, who explore LLMs in South Asian CM dialects through prompting experiments, and \citet{zhang-etal-2023-multilingual}, who assess LLMs' zero-shot performance on various CM tasks, including SA. Both studies find that LLMs need significant improvement on \textit{zero-shot} CM tasks but do not explore if LLM-generated data can help \textit{task-specific models} improve their training, despite sub-optimal LLM zero-shot performance. Notably, both studies found that GPT-3.5 \citep{brown2020language} shows superior performance among LLMs and do not evaluate the more advanced GPT-4 \citep{openai2024gpt4}. Our research builds on their findings by using GPT-4 for data generation and fine-tuning task-specific models in addition to evaluating zero-shot performance.

In contrast to the findings of \citet{yong-etal-2023-prompting} and \citet{zhang-etal-2023-multilingual}, \citet{whitehouse-etal-2023-llm} report improvements using GPT-4 for data augmentation in cross-lingual commonsense reasoning tasks. While cross-lingual tasks involve separate languages, code-mixed tasks involve language switching within sentences. Nonetheless, the success reported by \citet{whitehouse-etal-2023-llm} supports the feasibility of our approach.

\section{Methods}

%In this section, we outline our approach, which consists of two main steps: 1) data augmentation through few-shot prompting and 2) fine-tuning a multilingual PLM for sentiment classification.

%\subsection{Data Augmentation}
%Our work demonstrates the capability of GPT-4 for data augmentation in CM generation task and introduces a optimized sentiment analysis system for CM social media.

In this section, we introduce our data, the synthetic generation process, and our fine-tuning methods.

\subsection{Natural Data} \label{sec:data}

We conducted experiments using two human-labeled datasets which we call our natural data. The first is the Spanish-English SA dataset from the LinCE Benchmark \citep{aguilar-etal-2020-lince}, containing 18,789 CM tweets with code-mixing between English and Spanish. The second dataset is the Malayalam-English SA dataset from the MalayalamMixSentiment dataset \citep{chakravarthi-etal-2020-sentiment}, containing 5,452 CM YouTube movie review comments with code-mixing between English and Malayalam, a low-resource Dravidian language. The mean sentence lengths for both datasets are shown in Table \ref{table:datastatistics}.

Both datasets feature colloquial CM social media comments with diverse code-mixing patterns, presenting significant challenges to NLP models. They include sentiment categories: \textit{Positive}, \textit{Negative}, or \textit{Neutral}. For preprocessing, we filtered out comments labeled "non-Malayalam" or "unknown" from the Malayalam-English dataset and adjusted the data splits. Both datasets were cleaned to remove empty strings, hashtags, URLs, and symbols, with emojis replaced by English descriptions using the emoji library.\footnote{\url{https://pypi.org/project/emoji/}}

\begin{comment}
The task is to classify a social media comment into "Positive," "Negative," or "Neutral" based on its implicit sentiment.

%derived from SentiMix 2020 Spanish-English dataset\footnote{Spanish-English data: \url{https://zenodo.org/records/3974927#.XyxAZCgzZPZ}} 

Our experiments use the LinCE Benchmark's Spanish-English SA dataset \citep{aguilar-etal-2020-lince} consisting of 18,789 CM tweets, a subset of the Spanish-English SA dataset consisting of 3,000 CM tweets, and the MalayalamMixSentiment dataset \citep{chakravarthi-etal-2020-sentiment} consisting of 5,452 CM Youtube movie review comments. Both datasets include colloquial CM social media comments which reflect natural CM and pose a challenge to NLP models. Furthermore, they reflect diverse code-mixing patterns, as Spanish is a relatively high-resource Romance language and Malayalam is a low-resource Dravidian language.

%While we use the Spanish-English dataset (train: 12194, dev: 1859, test: 4736) for a majority of our experiments due to language comprehensibility, we test our system's transferability and generalizability with extremely scarce Malayalam-English data (train: 3452, dev: 1000, test: 1000).

For Malayalam-English, we filter out data points with labels "non-malayalam" or "unknown," and resplit the data according to the new size. We preprocess both datasets to remove empty strings, hash symbols, URLs, and at signs. We replace emojis with English descriptions using the emoji library\footnote{\url{https://pypi.org/project/emoji/}}.
\end{comment}

\begin{table}[ht]
\centering
\renewcommand{\arraystretch}{1.3} % Adjust row height
\setlength{\tabcolsep}{6pt} % Adjust column spacing
\scalebox{0.9}{ % Scale down the table
\begin{tabular}{p{0.23\columnwidth}| p{0.19\columnwidth} p{0.21\columnwidth} p{0.25\columnwidth}}
\hline
\textbf{Language} & \textbf{Natural} & \textbf{LLM-Generated} & \textbf{Random Translation}\\
\hline
Sp-En           & 13.0 \textpm 7.4 & 14.7 \textpm 4.0 & 23.1 \textpm 30.4 \\ 
\hline
Ma-En   & 8.2 \textpm 3.1 & 8.4 \textpm 1.7 & N/A \\  % Add N/A to the empty cell
\hline
\end{tabular}
}
\caption{Mean sentence length and standard deviation, in words, of natural and synthetic data for each Spanish-English (Sp-En) and Malayalam-English (Ma-En).}
\label{table:datastatistics}
\end{table}

\subsection{Data Augmentation Methods}

Our primary data augmentation method involves prompting LLM with task demonstrations to generate synthetic CM training data. As a secondary method to use for comparison, we implement the more traditional technique of translating monolingual sentences into CM. 

\subsubsection{LLM Prompting}
We use GPT-4 as our LLM, as many past studies \citep{whitehouse-etal-2023-llm, yong-etal-2023-prompting, zhang-etal-2023-multilingual} have found high CM performance in GPT-based models. We construct instructions for GPT-4 based on previously successful CM generation prompts \citep{whitehouse-etal-2023-llm, yong-etal-2023-prompting} and empirical observations of the data. Additionally, we provide task demonstrations randomly sampled from the natural pre-processed training dataset, which may again appear in the SA fine-tuning phase, with an equal amount of demonstrations for each class. Since LLM requires few task demonstrations, this data augmentation approach is not contingent on having a large dataset, and synthetic data generation utilized 15 to 50 examples. The prompt refinement process, our final prompt, and data generation implementation details can be viewed in Appendix \ref{sec:datagen}.

% Shown in Figure \ref{fig:prompt}, 

\begin{comment}
\begin{figure*}[t]
  \includegraphics[width=\textwidth]{prompt}
  \caption{Prompt refinement process (highlighted in yellow), GPT-4 responses (with unfavorable segments highlighted in red), and final prompt and response (right).}
  \label{fig:prompt}
\end{figure*}
\end{comment}

%Figure \ref{fig:prompt} displays examples of our prompts and LLM responses during the prompt refinement process, during which we iteratively adjusted our prompt based on errors the LLM made during preliminary experiments. Our final prompt, which we used to generate our synthetic datasets, is displayed on the right column of the table. We gave an equal number of task demonstrations for each class, and we adjusted the total number of task demonstrations given based on preliminary experimentation. Since LLM requires few task demonstrations, this data augmentation approach is not contingent on having a large dataset, and synthetic data can be generated based on only 15-500 task demonstrations.

Our final synthetic data sizes were \textasciitilde53,000 in Spanish-English and \textasciitilde24,000 in Malayalam-English. Shown in Table \ref{table:datastatistics}, LLM-generated sentences effectively resembled natural CM sentences in mean sentence length. However, LLM-generated sentences tended to vary less in sentence length, indicated by consistently lower standard deviation values.
 
%\section{Experimental Setup} \label{experimentalsetup}

%This section describes data augmentation implementation details and fine-tuning details.
% test split 3542,1000,1000. temperature 0.6. between \{15, 50, 150, 500\}

\subsubsection{Random Translation}\label{sec:random}
Our secondary technique, Random Translation, converts a monolingual SA corpus into a CM SA corpus using machine translation. Similar to \citet{li2022language,tareq,tarunesh-etal-2021-machine}, we used Stanford's Sentiment140 dataset \citep{go2009sentiment} and SemEval's Sentiment Analysis in Twitter dataset \citep{rosenthal-etal-2017-semeval} as monolingual corpora and randomly translated parts of English tweets into Spanish through Marian NMT \citep{junczys-dowmunt-etal-2018-marian}. We did not use this technique for Malayalam-English due to the lack of reliable machine translation systems supporting Malayalam.

The resulting synthetic corpus consisted of 49,560 data samples. As shown in Table \ref{table:datastatistics}, the randomly translated data exhibited a significantly higher mean sentence length compared to LLM-generated synthetic data, due to constraints imposed by the statistics of the selected monolingual dataset. This highlights the limited flexibility of using pre-existing datasets for CM data augmentation.

%\textbf{Language} & \multicolumn{2}{c}{\textbf{Natural}} & \multicolumn{2}{c}{\textbf{LLM-Generated}} & \multicolumn{2}{c}{\textbf{Random Translation}} \\ \hline
                   % & \textit{Mean} & \textit{STD}  & \textit{Mean} & \textit{STD}  & \textit{Mean} & \textit{STD}  \\ \hline
                    
\subsection{Fine-tuning Sentiment Analysis} \label{fine-tuning}
%Following the procedure of \citet{li-murray-2023-zero, whitehouse-etal-2023-llm,tareq}, We validate the effectiveness of data augmentation through fine-tuning multilingual pre-trained language models (PLMs) on synthetic data.

 We fine-tuned multilingual BERT (mBERT), which was most commonly used in past benchmarks \citep{chakravarthi-etal-2020-sentiment,aguilar-etal-2020-lince}, and XLM-T, which is a XLM-R \citep{conneau2020unsupervised} model pre-trained on millions of social media tweets from over thirty languages including Spanish and Malayalam. For each language, we trained both models on three datasets: only natural data, only synthetic data, and a combined dataset of natural and synthetic data. We also introduced a lower-resource experimental setup for Spanish-English, where we reduced the natural data to a 3,000-sample subset to align with \citet{li2022language}. Table \ref{table:datas} summarize the data sizes used in the full Spanish-English, subset of Spanish-English, and Malayalam-English experimental setups. For the full 12.2k Spanish-English data setup, we repeated experiments using both LLM-generated and randomly-translated synthetic data to compare the two techniques. In all, we hypothesized that training on both natural and synthetic data would lead to the highest performance, as it benefited from both natural data, which had a similar distribution and style as the natural test data, and synthetic data, which increased the number of examples for models to learn CM features.

 %While we initially experimented with multilingual BERT per past benchmarks and data augmentation work \citep{chakravarthi-etal-2020-sentiment,aguilar-etal-2020-lince}, we found higher results with XLM-T, so we used XLM-T for the rest of our experiments. Nonetheless, we include performance with mBERT in Section \ref{sec:results} to compare with other baselines and demonstrate model generalizability. 

\begin{table}[ht]
\centering
\renewcommand{\arraystretch}{1.2}
\setlength{\tabcolsep}{6pt}
\begin{tabular}{p{0.2\columnwidth} p{0.15\columnwidth} p{0.15\columnwidth} p{0.1\columnwidth} p{0.1\columnwidth}}
\hline
\textbf{Language} & \multicolumn{2}{c}{\textbf{Train}} & \textbf{Val} & \textbf{Test} \\ \hline
                    & \textit{Natural} & \textit{Synthetic} &  &  \\ \hline
Sp-En           & 12,194  & 50,000  & 1,859 & 4,736 \\ 
\hline
Sp-En & 3,000   & 50,000  & 1,859 & 4,736 \\ \hline
Ma-En   & 3,452   & 15,000  & 1,000 & 1,000 \\ 
\hline
%Sp-En & 12,194   & 49,560  & 1,859 & 4,736 \\ \hline
%Sp-En & 3,000   & 49,560  & 1,859 & 4,736 \\ \hline
\end{tabular}
\caption{Training, validation, and test data sizes for each round of experiments. Each row included training on natural data, synthetic data, and the combined (natural + synthetic) data, repeated for mBERT and XLM-T across different types of synthetic data in Spanish-English (Sp-En) and Malayalam-English (Ma-En).} \label{table:datas}
\end{table}

In all Spanish-English experiments, when training on a combination of synthetic and natural data, we adopted the gradual fine-tuning method proposed by \citet{xu-etal-2021-gradual} and applied to CM data augmentation by \citet{li2022language}. Treating the synthetic CM data as out-of-domain data, we fine-tuned the model for five stages, gradually decreasing the amount of synthetic data from 50,000 to \{25000, 15000, 5000, 0\} for subsequent training stages while keeping natural data constant. As a result, the model gradually fit closer to natural data, which it would be tested on. In Malayalam-English, we retained one stage of training due to higher performance after preliminary experimentation. Fine-tuning hyperparameters and the impact of gradual fine-tuning are included in Appendix \ref{sec:finetunedetails} and Appendix \ref{sec:Gradual}, respectively.

\section{Results} \label{sec:results}

This section evaluates overall model performance and then quantifies relative percent improvements contributed by data augmentation. %XLM-T and mBERT performance on the full Spanish-English dataset, the subset Spanish-English dataset, and the Malayalam-English datasets and compares them to prior work.
%which is sensitive to error, used for imbalanced datasets, and the most common metric used for classification both in this domain () and in general (). We evaluate on human-made test data, which are unseen to both LLM and Sentiment Classifier. For Spanish-English, we use the LinCE benchmark evaluation system. For Malayalam-English, we use our 1000 split test data.
\begin{table*}[ht]
\centering
\scalebox{1}{
\begin{tabular}{p{0.2\textwidth} p{0.1\textwidth} p{0.1\textwidth} p{0.1\textwidth} p{0.13\textwidth} p{0.15\textwidth}}
    \toprule

    \textbf{Method} & \textbf{Model} & \textbf{Natural Data} & \textbf{Synthetic Data} & \textbf{Spanish-English F1} & \textbf{Malayalam-English F1}\\

    \midrule
    
    Zero-shot & GPT-4 & & & 0.546 & 0.524\\

    No Training & mBERT & & & 0.045 & 0.131\\

    No Training & XLM-T & & & 0.543 & 0.354\\

    \hline
    
    Dataset Baseline & mBERT & \checkmark & & 0.564 & 0.750\\

    Our Baseline & XLM-T & \checkmark & & 0.588 & \textbf{0.843}\\

    \hline
    Random Translation & XLM-T & & \checkmark & 0.491 & \\
    LLM-Generated & XLM-T & & \checkmark & 0.544 & 0.595\\

    \hline
    Random Translation & XLM-T & \checkmark & \checkmark & 0.563 & \\
    LLM-Generated & XLM-T & \checkmark & \checkmark & \textbf{0.603} & 0.763\\

    \hline
    Top Score & & & & 0.622 & 0.804\\
    
  \end{tabular}
}
\caption{Summary of weighted F1 scores on the full 12k Spanish-English and 3.5k Malayalam-English datasets with comparisons to other baselines. Scores in bold indicate our highest performance on each dataset. The top score for Spanish-English is anonymous on the LinCE benchmark, and the top score for Malayalam-English is \citet{bai2021}.} \label{tab:General}
\end{table*}

\subsection{Overall Performance}

Table \ref{tab:General} presents the overall F1 scores achieved for the Spanish-English and Malayalam-English CM SA datasets in the full 12.2k and 3.5k data setup, respectively, compared to zero-shot scores, baseline scores, and current benchmarks. All Spanish-English models were evaluated using the same test dataset as the LinCE benchmark. However, the Malayalam-English models used adjusted train-test splits in comparison to benchmarks, due to the removal of extraneous labels (see Section \ref{sec:data}). %while Table \ref{tab:comparison} focuses on the contribution of data augmentation and the relative percent improvements. %We use the LinCE Benchmark's dataset and website\footnote{\url{https://ritual.uh.edu/lince/home#}} to evaluate all Spanish-English methods since the labels of the Spanish-English test data are not publicly released. %Additional information about precision and recall scores are included in Appendix \ref{section:precision}.

%zero-shot
%\subsubsection{Overall Performance}\label{sec:overall}
%In Table \ref{tab:General} we provide zero-shot scores, synthetic data scores, and natural baseline scores on the Spanish-English CM SA task (rows 1-7) for context.

\subsubsection{Baselines}
To provide reference points, GPT-4, mBERT, and XLM-T were evaluated using a zero-shot approach, where no additional training or fine-tuning was applied. For GPT-4, we generated predictions by providing a prompt with no examples and parsing the generated outputs directly as the model's predictions. For mBERT and XLM-T, we loaded in the pre-trained models with an extra classification layer and proceeded directly to evaluation without further training. Results are shown in the first section of Table \ref{tab:General}.

Our zero-shot analysis reveals three main findings. First, consistent with \citet{zhang-etal-2023-multilingual}, large language models like GPT-4 are still not sufficiently adept for zero-shot tasks like Spanish-English and Malayalam-English sentiment analysis, as they perform below dataset benchmarks \citep{aguilar-etal-2020-lince, chakravarthi-etal-2020-sentiment}. However, GPT-4's zero-shot performance on Malayalam-English is still surprisingly high considering the language is low-resource. Second, the size of an LLM does not necessarily equate to better performance. XLM-T, with its task-specific pre-training on code-mixed data from Common Crawl and Twitter \citep{li2022language}, demonstrates that a smaller, specialized model can be nearly as effective as a much larger general-purpose model in Spanish-English, aligning with \citet{zhang-etal-2023-multilingual}. Lastly, XLM-T shows a significant zero-shot performance boost over mBERT for both Spanish-English and Malayalam-English, demonstrating the importance of task-specific pre-training.

The second section of Table \ref{tab:General} shows results after fine-tuning XLM-T on the full natural data. XLM-T consistently outperforms mBERT in both languages, similar to its zero-shot performance. Our Spanish-English baseline with XLM-T surpasses the LinCE Organizers' baseline using mBERT, and our Malayalam-English baseline achieves the highest score on this dataset, exceeding the previous top score by \citet{bai2021}.

%The second section of Table \ref{tab:General} displays results after fine-tuning XLM-T on natural data. Like its zero-shot performance, XLM-T achieves higher results than mBERT when fine-tuned on the same natural data for either language. Our Spanish-English baseline, which uses XLM-T, improves on the LinCE Organizers' baseline, which uses mBERT under the same conditions. Furthermore, our Malayalam-English baseline is extremely effective and achieves the highest score overall on this dataset, outperforming the previous highest score \citet{bai2021} by xx \%.%\footnote{It is important to note that past benchmarks on the Malayalam-English dataset perform 5-label classification, while our focus is on 3 labels, so our datasets slightly deviate.}

%We use XLM-T fine-tuned on the full natural sets as our baselines for both languages.

 %However, since XLM-T already performs well on this task, it makes sense that the relative improvements made by training and data augmentation may be less substantial.

%ows 6 and 11 in the Spanish-English dataset are from the LinCE Benchmark \cite{aguilar-etal-2020-lince}. It is important to note that past benchmarks on the Malayalam-English dataset perform 5-label classification, while our focus is on 3 labels, so our datasets slightly deviate. 

 \subsubsection{Performance with Synthetic Data}

The third and fourth sections of Table \ref{tab:General} display results when fine-tuning XLM-T on solely synthetic data and on a combination of natural and synthetic data, respectively.

When fine-tuning XLM-T on solely synthetic Spanish-English data, LLM-generated data slightly improves performance compared to no training, whereas randomly-translated data decrease performance below zero-shot levels. 

Combining random-translated data with the full natural Spanish-English data similarly degrades performance relative to our baseline, highlighting its less effective representation of code-mixing. On the other hand, combining natural and LLM-generated synthetic data yields our highest Spanish-English score of 0.603 F1, ranking third on the LinCE benchmark. This demonstrates that LLM-generated data can mitigate overfitting and enhance task-specific model performance beyond LLM's own zero-shot capabilities in Spanish-English.

For Malayalam-English, training on either synthetic or natural data significantly improves performance compared to zero-shot results. LLM-generated synthetic data nearly double XLM-T's performance, and natural data more than double it, achieving higher scores than Spanish-English. Training with both natural and synthetic data averages their individual performances, suggesting that there exists a performance threshold past which synthetic data can no longer help. Nonetheless, the combination surpasses the dataset benchmark \citep{chakravarthi-etal-2020-sentiment}.

\subsection{Contribution of Data Augmentation}\label{sec:contribution}

% 1) compare the three big percent changes
% 2) compare each small one
% 3) mention tradeoffs with adding human data to 3000
% 3) mention that llm augmentation decreased results but sitll beat li and murray overall
% 4) mention xlm-t vs mbert discrepancy and how it impacts contribution

\begin{table*}
\centering
%\begin{tabular}{@{}l *{4}{S} *{4}{S}@{}}
\begin{tabular}{llllll} %{|l|l|l|l|l|l|l|l|l|l|}%
    \toprule

 \textbf{Dataset} & \textbf{Method} & \textbf{Model} & \textbf{Baseline} & \textbf{+Synthetic} & \textbf{\% Change}\\

    \midrule
    
    \multirow{3}{*}{Full Spanish-English$_{12.2k}$} 
    
    & LLM-Generated & XLM-T &  0.588 & 0.603 & 2.55\%\\

    & LLM-Generated & mBERT & 0.503 & 0.533 & \textbf{5.96\%}\\

    & Random Translation & XLM-T& 0.588 & 0.491 & -16.5\%\\

    & Random Translation & mBERT & 0.503 & 0.512 & 1.79\%\\

    \hline

    \multirow{3}{*}{Subset of Spanish-English$_{3k}$}
    
    &LLM-Generated & XLM-T & 0.547 & 0.598 & \textbf{9.32\%}\\

    &LLM-Generated & mBERT & 0.487 & 0.526 & 8.01\%\\
    
    %\hline
    & \citet{li2022language} & XLM-T  & 0.649 & 0.660 & 1.68\%\\    
   & \citet{li2022language} & mBERT& 0.495 & 0.506 & 2.12\%\\

    \hline

    \multirow{2}{*}{Malayalam-English$_{3.5k}$}
    
     & LLM-Generated & XLM-T  & 0.843 & 0.763 & -9.84\%\\

     & LLM-Generated & mBERT  & 0.737 & 0.745 & 1.09\%\\

    & \citet{li2022language} & mBERT & 0.670 & 0.722 & \textbf{7.73}\%
    %hline
    
    %hline

  \end{tabular}
  \caption{A comparison of relative percent improvements achieved by different data augmentation methods on our three datasets for XLM-T and mBERT, with the largest improvements highlighted in bold. F1 scores are also provided from fine-tuning on natural data and on a combination of natural and synthetic data.} \label{tab:comparison}

\end{table*}

Table \ref{tab:comparison} displays the relative improvements from data augmentation techniques on the three data setups: the full Spanish-English dataset, the subset of the Spanish-English dataset, and the Malayalam-English dataset. Unlike absolute scores, which can vary with training conditions, percent improvements provide a consistent measure for comparing models trained with and without synthetic data.

\subsubsection{Full Spanish-English Dataset}
The contrast in relative improvements between the LLM-Generated technique and the Random Translation technique, which are shown in the first section of Table \ref{tab:comparison}, can be attributed to two factors: First, the monolingual corpora used for Random Translation did not closely match the distribution of natural CM data, and second, the code-switching points in the synthetic data were randomly generated. Since LLM-generated data did not experience the same performance losses, it mitigated these issues by producing sentences that more accurately reflected natural data distributions and incorporated intentional code-switching rather than random occurrences.

\subsubsection{Subset of Spanish-English Dataset} \label{sec:subset}
In the subset of the Spanish-English dataset, where the training set was reduced to 3,000 samples, LLM-generated data showed a more substantial improvement for both models than on the full Spanish-English dataset, displayed in the second section of Table \ref{tab:comparison}. These improvements outperformed the results obtained by \citet{li2022language}, indicating that LLM-generated data samples are particularly effective in a Spanish-English low-resource setting.
\subsubsection{Malayalam-English Dataset}
Displayed in the third section of Table \ref{tab:comparison}, the high baseline accuracy of XLM-T in Malayalam-English led to a performance drop with synthetic data, while mBERT's performance improved slightly. In comparison, \citet{li2022language} cite large improvements using their language-agnostic method, which reduces the focus on Malayalam's particular language features and emphasizes learning CM patterns. Nonetheless, this method also improves on a lower baseline score. These disparities suggest that the utility of synthetic data may diminish when the model’s baseline performance is already high.

%The third section of Table \ref{tab:comparison} displays results on Malayalam-English, where we find decreases in results in comparison to \citet{li2022murray}. We hypothesize that this is due to Malayalam-English's high baseline results and its low-resource nature. We elaborate on this hypothesis in Section \ref{sec:analysis}. While \citet{li2022murray}'s method is more effective in improving results, it is notable that our combined natural and synthetic data score still beat its score despite having a low percent improvement.

\subsubsection{Cross-Dataset Analysis}
Across all datasets, synthetic data generally enhances performance up to a certain threshold. Models with lower initial baselines, such as those trained on the limited Spanish-English subset, show greater percent improvements with synthetic data, reaching almost the same performance as models with quadruple the amount of natural data. This performance stability suggests that LLM-powered data can effectively boost performance for relatively small datasets. Conversely, models with high initial baselines, like XLM-T in Malayalam-English, may experience a decrease in accuracy when synthetic data samples are added, as synthetic data maintain performance at a similar threshold.

Overall, LLM-powered data augmentation proves effective in improving five of six models for CM SA, with our Spanish-English system achieving a notable 9.32\% relative percent improvement, surpassing other methods such as \citet{li2022language} under similar conditions.

\section{Analysis}\label{sec:analysis}

This section details results from human evaluation, subsequent empirical data analysis, and discussion about the trade-offs of generating synthetic data.

\subsection{Human Evaluation}\label{sec:humaneval}

To gain insight on the quality of LLM-generated data, we asked native speakers to evaluate Spanish-English and Malayalam-English sentences from both the original dataset and the LLM-generated dataset on the grounds of \textit{Code-Mixing Naturalness,} \textit{Label Accuracy,} and if the sentences are \textit{Human or Machine-Generated.} 400 Malayalam-English sentences were labeled by one annotator, and 200 Spanish-English sentences were labeled by two annotators, all of whom were balanced bilinguals with C1-C2 proficiency in the languages they annotated, according to the Common European Framework of Reference for Languages (CEFR). Detailed instructions for evaluators and descriptions of each label are elaborated in Appendix \ref{sec:human}. In this study, our human evaluation was constrained due to limited resources. While this is a limitation, it is worth noting that other studies, such as \citet{whitehouse-etal-2023-llm}, have worked with even smaller sample sizes.

\begin{figure}
  \includegraphics[width=0.95\columnwidth]{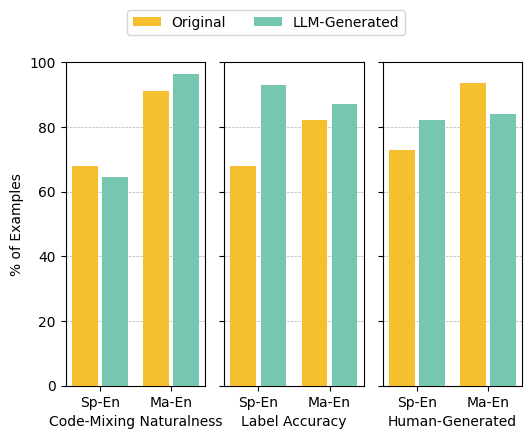}
  \caption{Human evaluation on Spanish-English and Malayalam-English sentences from the original datasets and the LLM-Generated datasets.}
  \label{fig:agree}
\end{figure}

%The annotators are asked to choose from "natural, like something people would actually type/say," "a bit strange/could be improved," and "unnatural/needs to be rewritten," (2) 

As shown in the first graph of Figure \ref{fig:agree}, annotators rated LLM-generated sentences similarly to human-generated sentences in terms of naturalness for both datasets. This suggests that LLM-generated sentences did not appear unnatural when compared to human sentences. Notably, our Malayalam-English annotator labeled 5.5\% more synthetic sentences as natural compared to human sentences. Since we define CM naturalness as the fluency of a sentence such that it can be recognized and accepted as authentic CM in real-life contexts, this finding indicates that, despite the differences in appearance between LLM-generated and natural data, both forms may be perceived as valid representations of CM in the real world. Furthermore, while there is a slight increase in the rating of synthetic sentences in Malayalam-English, the difference is relatively small and may not represent a significant divergence between LLM-generated and human sentences in terms of perceived naturalness.
 
%Native speakers found that LLM-generated sentences are indistinguishable from human sentences. Of 400 Malayalam-English sentences, 6 were labeled as "unnatural" and all 6 were actually from the human-made set. 15 were labeled as somewhat unnatural, with 4 coming from the synthetic data and 11 coming from the human data. For Spanish-English, slightly more synthetic sentences were labeled as unnatural. Of 200 sentences, 19 were labeled as unnatural, and 12 were from synthetic while 7 were from human. 51 were labeled as somewhat unnatural, 26 from synthetic and 25 from human. This indicates that the unnatural ratings were more due to the constant and not due to the actual unnaturalness. Nonetheless, naturalness scales were pretty well distributed between human and synthetic data, indicating the natural quality of the synthetic data. 

Consistent across both datasets, LLM-generated data exhibited significantly higher sentiment label accuracy compared to human-generated data, shown in the second graph of Figure \ref{fig:agree}. This finding suggests that LLM-generated samples are less ambiguous, likely because we explicitly prompt GPT-4 to generate sentences for the sentiment analysis task. In contrast, real-world social media tweets, created without this directive, may exhibit greater semantic variability. These results highlight potential label ambiguity issues in the original datasets, particularly for Spanish-English, and demonstrate the utility of synthetic sentences to mitigate these issues by providing clearer examples during training. However, less ambiguous synthetic data may also lead to models that are less robust to natural complexities in human expression.
% Nonetheless, clearly labeled examples are essential because they help prevent models from developing incorrect or biased interpretations of words or meanings in sentences due to labeling inaccuracies.

When predicting whether a sentence was human- or machine-generated, annotators faced significant challenges in distinguishing between LLM-generated and human sentences, shown in Figure \ref{fig:agree}. For Spanish-English, annotators mistakenly identified more LLM-generated sentences as human-produced than actual human sentences. In Malayalam-English, while annotators more accurately identified human sentences, a substantial margin of error persisted. Consequently, even though annotators tended to rate certain groups with higher naturalness or label accuracy, they lacked a clear understanding and identifiable cues indicating the sentences' original sources.

Ultimately, inter-annotator agreement was low for Spanish-English ($\kappa < 0.3$). While our findings offer a qualitative perspective to the quantitative fine-tuning results, we encourage more comprehensive studies dedicated to human evaluation in the future.

\subsection{Empirical Data Analysis} \label{empirical}

When observing natural and synthetic data, we focus on explaining two questions: (1) Why did the Malayalam-English baseline perform better than Spanish-English despite less training data? (2) Why did synthetic data improve Spanish-English performance while decreasing Malayalam-English performance in XLM-T? We find that the challenges in the dataset, task, and the training background of LLM best answer these questions.

%The Spanish-English dataset is more challenging, and because of that, it gets a lower baseline performance. The higher baseline performance in Malayalam-English and the differing similarity in LLM-generated sentences helps to explain why synthetic data improved Spanish-English more than Malayalam-English.

\subsubsection{Dataset Challenge}

Aligning with the results of human evaluation, we found significant label ambiguity in the human-labeled Spanish-English dataset due to both the inherent ambivalence of human speech and the various interpretations that can be made by human annotators.

\begin{table}
\centering
%\begin{tabular}{@{}l *{4}{S} *{4}{S}@{}}
\scalebox{0.75}{
\begin{tabular}{p{0.8\columnwidth} p{0.12\columnwidth} p{0.22\columnwidth}} %{|l|l|l|l|l|l|l|l|l|l|}%
    \toprule

 \textbf{Sentences}
& \textbf{Label}
& \textbf{Prediction}\\
\midrule
 Happy Friday \#elvacilondelaGatita
&neutral
& positive\\
 \#elvacilondelagatita \#quotes \#friday
&positive
& neutral\\

    \toprule

 \textbf{Sentences}
& \textbf{Label}
& \textbf{Correction}\\
\midrule
Get your outfit now! Escoge tus prendas favoritas y haz tu pedido Blusa morada \$20.00 \#ilovesalhuaclothing $\heartsuit$ & neutral & positive\\
Como me encabrona enterarme de quien se va en The Bachelor sin haber visto el episodio Angry Face & positive & negative\\

    \bottomrule
  \end{tabular}
  }
  \caption{Examples of sentences from the natural Spanish-English dataset, including their true labels, XLM-T's predicted labels, and the proposed corrections by human evaluators. Translations of the CM sentences into English are provided in Table \ref{table:ambtrans} in Appendix \ref{sec:translations}.} \label{table:amb}
\end{table}

In Table \ref{table:amb}, the first two examples highlight annotation ambiguity. Despite conveying similar ideas of anticipating Friday and listening to the Hispanic radio morning show "El Vacilón de la Gatita," they are labeled differently. Notably, the use of "Happy" in the first sentence seems to imply a positive sentiment but is labeled as neutral.

The subsequent examples illustrate disagreements between human evaluators and true labels. One example, a clothing ad with a seemingly positive connotation, could be interpreted as neutral due to its advertising context. Conversely, the second example, discussing hearing a spoiler for "The Bachelor," seems to clearly warrant a negative rather than positive label.

In contrast, the Malayalam-English dataset contains cleaner, more consistently phrased examples. A significant portion of negatively labeled sentences include the word "Dislike," simplifying the sentiment analysis task. This consistency likely contributes to Malayalam-English's high performance compared to Spanish-English. We provide further analysis of the challenges of CM sentiment analysis in Appendix \ref{section:laugh}.

%In Table \ref{table:amb}, the first two examples demonstrate annotation ambiguity. While both sentences convey the same idea (it is Friday and time to listen to a specific podcast), they are labeled differently. Furthermore, these labels seem reversed, as the first sentence uses "Happy" which carries a more positive connotation yet is labeled "neutral", reflecting human labeling subjectivity.

%The second two examples show sentences where human evaluator collectively disagreed with true labels. The first of the two examples is a positively-worded clothing ad, which can be viewed as positive due to the use of the word "love" or neutral because ads can also be perceived with a negative connotation. On the other hand, the second of the two examples shows a more explicit error that human evaluators caught. The sentence translates to "How did I find out who is leaving The Bachelor without having seen the episode Angry Face," which seems to merit a negative or neutral label rather than positive.

%On the other hand, we observed that Malayalam-English data is cleaner and contains more consistently phrased examples. For example, there are 49\% of sentences with a negative label in the natural training set and 43\% in the test set include the word "Dislike," which is a commonly said English word in Malayalam-English, especially on Youtube. This data inherently simplifies the task, providing context for Malayalam-English's high performance compared to Spanish-English.

\subsubsection{Data Parallels}

We discovered many parallels between natural and synthetic Spanish-English data both semantically and syntactically. Shown in Table \ref{tab:parallel}, both natural and synthetic sentences discuss common ideas, such as replaying a song, and use Spanish-English hybrid words like "textear." LLM's ability to adapt to the topics discussed in the Spanish-English data and to capture these CM nuances supports the high performance gains synthetic data provide.

\begin{table}[htbp]
\centering
%\begin{tabular}{@{}l *{4}{S} *{4}{S}@{}}
\scalebox{0.67}{\begin{tabular}{p{0.7\columnwidth}p{0.7\columnwidth}} %{|l|l|l|l|l|l|l|l|l|l|}%
    \toprule
      \textbf{Natural} & \textbf{Synthetic}  \\
      \midrule

    \textcolor{darkgray}{Something came up} \textcolor{purple}{algo surgió un problema} \textcolor{darkgray}{\textbf{sorry something came up} and I cann't make it to the party} & \textcolor{darkgray}{Can’t believe I got stood up...} \textcolor{purple}{Mi} \textcolor{darkgray}{date} \textcolor{purple}{dijo} \textcolor{darkgray}{\textbf{"sorry, something came up"} like for real?!} \\

    \midrule
    
    \textcolor{purple}{Ojos verdes on} \textbf{replayyy}  & This song \textcolor{purple}{me tiene} in my feels, \textbf{replay} x100 \\

    \midrule
    
    \textcolor{purple}{Estaba pensando en} \textbf{textear}\textcolor{purple}{le a mi hermana y al minuto me llega un mensaje de ella} \#sisterlyconnection & \textcolor{purple}{Cuando te voy a} \textbf{textear} \textcolor{purple}{y apareces} typing, call it telepathy or just \textcolor{purple}{buena onda} \\
    
   %\midrule
   % \textcolor{purple}{Ke me compre un carrito pa irme con mis} friends and party lol & Dude, \textcolor{purple}{tu} ex estaba en the party, y yo tipo AWKWARD. \\

\midrule

    \textcolor{purple}{Se me olvidaron todos los} \textbf{passwords} \textcolor{purple}{del Hospital \textbf{y no podia entrar a} ningun lado \#PerksDeLosFinales} & UGH, \textcolor{purple}{olvidé mi} \textbf{password} again \textcolor{purple}{\textbf{y no puedo entrar a} mi cuenta...} FML \\

\midrule

    %\textcolor{purple}{Pronunciación: Muchas veces no nos entiendes porq agregamos E antes de la S en} School, Spanish, \textcolor{purple}{no es} eschool, espanish \textcolor{purple}{es} ssschool sspanish & \textcolor{purple}{Creo que} I finally got the hang of \textcolor{purple}{esto de} code-switching, it's kinda fun!\\

    \textcolor{darkgray}{Deslike} \textcolor{olive}{adicha ella punnara makkalkum nanni} & \textcolor{darkgray}{Plot had potential, but execution} \textcolor{olive}{polilla} \color{darkgray}{, disappointed.} \\
    \bottomrule
  \end{tabular}}
  \caption{Comparisons of natural and synthetic sentences in Spanish-English (red) and Malayalam-English (yellow). Overlapping words or phrases are highlighted in bold. Translations of the CM sentences into English are provided in in Table \ref{tab:paralleltrans} in Appendix \ref{sec:translations}.} \label{tab:parallel}
\end{table}

While the Spanish-English natural data frequently featured \textit{alternational} CM patterns, where sentences alternated between languages, Malayalam-English natural data primarily exhibited \textit{insertional} CM, where English words were occasionally inserted into predominantly Malayalam sentences. LLMs often generated alternational CM in Spanish-English and insertional CM with English as the dominant language in Malayalam-English. As a result, they improved performance in Spanish-English but did not align well with the Malayalam-English natural dataset, where Malayalam was the dominant language. Even though LLM-generated Malayalam-English data sounded natural according to human evaluators, it reflected a real-world insertional CM pattern not present in our particular human-labeled dataset. As a result, this discrepancy highlights the inherent complexity of CM tasks for ML models due to the diverse nature of CM cultural practices.

 A key challenge remains in controlling the type of CM—whether alternational or insertional—that LLMs produce. While LLMs handle alternation between English and Spanish with relative ease due to extensive training data, balancing languages like Malayalam and English remains a significant challenge. Consequently, the effectiveness of data augmentation is contingent not only the model's initial task performance but also the similarity between the CM patterns in natural and synthetic datasets.

\subsection{Trade-offs with Using Synthetic Data}\label{sec:cost}
While our research demonstrates that LLMs can effectively generate CM training data, the key question is why we should prefer LLM-generated data over human-labeled data.

Collecting high-quality natural CM data is resource-intensive, involving web scraping, human annotation, and rigorous quality control. For instance, to create the Spanish-English SA dataset, \citet{patwa-etal-2020-semeval} scraped CM data from Twitter, employed three Amazon Mechanical Turk\footnote{\url{https://www.mturk.com/}} workers to label 18,789 tweets, and conducted manual reviews to correct errors. The estimated cost for annotating these tweets was approximately \$3,054 USD, based on the minimum rate for Spanish-speaking workers.\footnote{Minimum rates for workers with premium qualifications are detailed here: \url{https://requester.mturk.com/pricing}} A detailed cost breakdown is available in Appendix \ref{section:cost}.

Comparing the baseline scores on the full Spanish-English dataset to the subset in Section \ref{sec:contribution}, adding \textasciitilde9,000 human-labeled sentences to a baseline of 3,000 resulted in a \textbf{7.49\%} improvement. According to the procedure above, the cost of these sentences was approximately \textbf{\$1,495 USD}, and the annotation process likely took several weeks.

In contrast, generating synthetic data using GPT-4 for both Spanish-English and Malayalam-English, including preliminary experiments, cost \$376.54 USD in total. Adding 50,000 synthetic sentences to the same baseline of 3,000 resulted in a \textbf{9.32\%} improvement. These sentences were generated in hours and cost only \textbf{\$37.92 USD}, making synthetic data generation 40 times cheaper than manual annotation of a corpus one-fifth the synthetic size.

While a larger volume of synthetic sentences is needed to achieve the same performance gains as a smaller set of human-labeled sentences, synthetic data generation is significantly more cost-effective and faster. Moreover, adding a large amount of synthetic data to natural data yields greater performance improvements (9.32\%) than adding a smaller set of human-labeled data (7.49\%).

\section{Conclusion and Future Work}

% THREE MAIN FINDINGS: 1) 60.3 THIRD PLACE 2) Malayalam-English new benchmark 3) 9% improvement on Spanish-English!!!!
To address CM data scarcity, we propose using few-shot prompting with LLMs to generate synthetic, labeled CM data for SA. We tested this approach by training mBERT and XLM-T on natural, synthetic, and combined datasets for Spanish-English and Malayalam-English. In Spanish-English, our method improved sentiment classification by 9.32\% for the 3k training setup and achieved third place on the LinCE benchmark for the 12k training setup. Human evaluations confirmed that our synthetic data closely mimic natural data and are indistinguishable from human-labeled examples. For Malayalam-English, our baseline system exceeded the highest published benchmark with an F1 score of 0.847, though further improvements with additional data were limited. Our findings indicate that LLM-generated synthetic data are most effective for enhancing models with low baseline performance, particularly when the languages are evenly represented as well as for resource-constrained scenarios. Overall, LLM-powered data augmentation offers a cost-effective alternative to human annotation, producing high-quality, natural-sounding sentences with minimal label ambiguity.

To improve performance in Malayalam-English, we intend to apply our observations of synthetic data to refine our LLM prompt and regenerate data. In addition, we aim to extend our research to encompass a broader range of LLMs and dialects, including those without English as a base and those primarily written in non-Latin scripts. Ultimately, our findings offer a promising avenue for CM data augmentation, and we encourage further exploration with LLMs in CM, an area which presents technical challenge and valuable social impact.

\section{Limitations}

The findings may not generalize across all types of data or tasks. While we find that results are generalizable across different PLMs such as mBERT and XLM-T and that LLMs typically generate natural-sounding sentences, the effectiveness of the data augmentation method may vary depending on the specific characteristics of the dataset, the resource level of the language, or the nature of the natural language processing task. Our experiments focused on Spanish-English and Malayalam-English for sentiment analysis, and we encourage future research to explore this method in other languages and tasks.

Furthermore, the effectiveness of this data augmentation method is limited by the baseline performance on natural data. If performance on natural data is already higher than the threshold synthetic data can raise results to, then further improvements are difficult to achieve. To mitigate this issue, an option is to regenerate synthetic data with an improved prompt, resulting in more natural synthetic data that can raise performance to an even higher threshold.

However, quickly quantifying the effectiveness of a prompt or strategy is challenging because it necessitates repeatedly generating large datasets and retraining models to measure performance improvements, which may become resource intensive if repeated numerous times. Furthermore, human evaluation was constrained to 200 and 400 data samples due to limited resources. In the future, developing a metric to quantify synthetic data quality without fine-tuning a separate model or using human evaluation would help streamline the development process and provide more direct insights.

%Another limitation is that it may be difficult to find LLMs trained on some extremely low-resource languages. While we find that GPT-4 shows impressive performance on CM variations with low-resource languages like Malayalam, it may not be the case for all other languages. This is a common issue faced in many NLP fields that is currently being worked on.
 
Notably, there are data augmentation methods for CM SA other than \citet{li2022language} and similar to our implementation of Random Translation, including \citet{tareq}, who convert a monolingual English corpus into Bangla-English using a word embedding algorithm, and \citet{ma-etal-2020-xlp}, who also randomly translate parts of a monolingual English corpus into Spanish-English. However, they either use different datasets, do not provide all baseline scores to be able to compare, do not detail their exact experiments, or do not release their code, so we were not able to directly compare our techniques with theirs.

\section{Ethics Statement}

 Like most data augmentation methods, LLM-powered synthetic data generation raises ethical concerns because of its potential to magnify biases within datasets. Since multilingual NLP and CM are interlaced with people's identities, cultures, and heritages, it is important that LLMs do not misrepresent peoples' cultures and languages in offensive or inaccurate ways. As a result, we acknowledge the importance of working alongside qualified CM experts and including speakers familiar with the languages in CM patterns in the research process. Before deploying models to the public, it is vital that generated data is verified and CM language models are thoroughly tested.

\section{References}

\bibliography{anthology,custom}
\bibliographystyle{acl_natbib}

\appendix

\section{Translations of Tables and Figures} \label{sec:translations}

\begin{figure}[htbp]
  \includegraphics[width=\columnwidth]{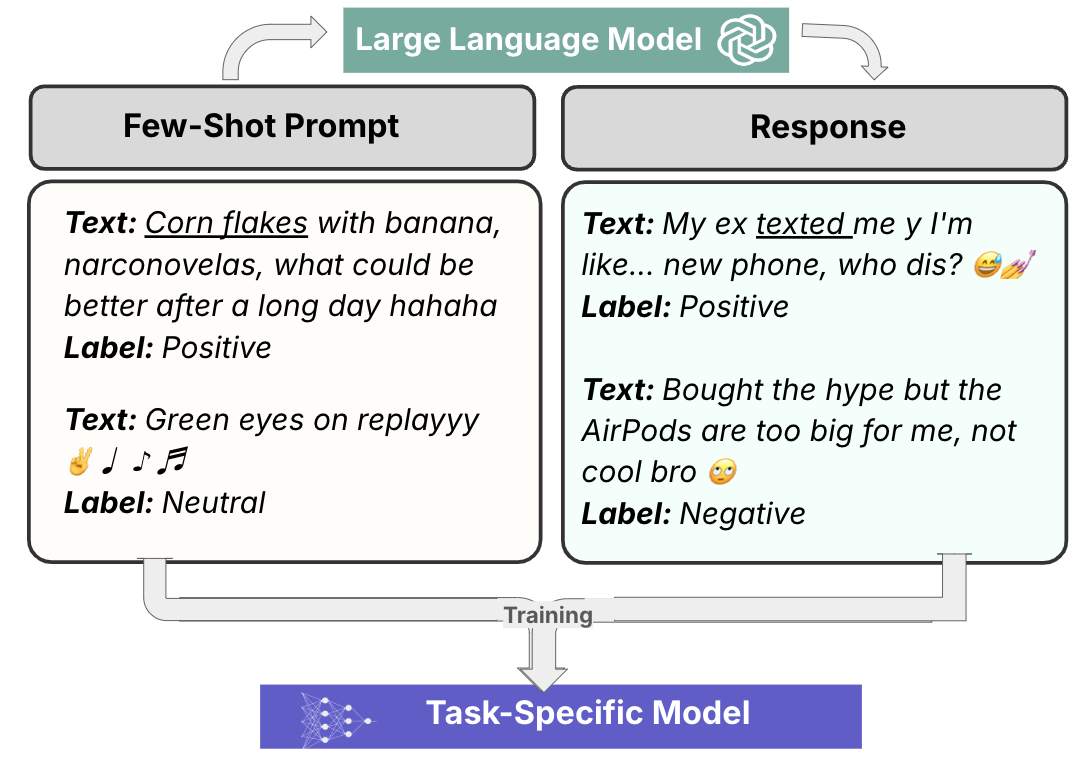}
  \caption{Overall system workflow with translated examples of Spanish-English CM tweets as natural data (left) and synthetic data (right). Underlined words represent Spanish-English hybrid words, examples of the complexities introduced by CM.}
  \label{fig:translationex}
\end{figure}

This section provides translations of the CM sentences used in Figure \ref{fig:examples} and in the tables in Section \ref{empirical}. Figure \ref{fig:translationex} is a translated version of Figure \ref{fig:examples}, Table \ref{table:ambtrans} is a translated version of Table \ref{table:amb}, and Table \ref{tab:paralleltrans} is a translated version of Table \ref{tab:parallel}.

\begin{table}
\centering
%\begin{tabular}{@{}l *{4}{S} *{4}{S}@{}}
\begin{tabular}{p{0.5\columnwidth} p{0.12\columnwidth} p{0.15\columnwidth}} %{|l|l|l|l|l|l|l|l|l|l|}%
    \toprule

 \textbf{Sentences}
& \textbf{Label}
& \textbf{Prediction}\\
\midrule
 Happy Friday \#thejokeoftheKitten
&neutral
& positive\\
 \#thejokeoftheKitten \#quotes \#friday
&positive
& neutral\\

    \toprule

 \textbf{Sentences}
& \textbf{Label}
& \textbf{Correction}\\
\midrule
Get your outfit now! Choose your favorite garments and place your order Purple blouse \$20.00 \#ilovesalhuaclothing $\heartsuit$ & neutral & positive\\
How I find out who's leaving on The Bachelor without having seen the episode Angry Face & positive & negative\\

    \bottomrule
  \end{tabular}
  \caption{Translated examples of sentences from the natural Spanish-English dataset, including their true labels, XLM-T's predicted labels, and the proposed corrections by human evaluators.} \label{table:ambtrans}
\end{table}

\begin{table}[htbp]
\centering
%\begin{tabular}{@{}l *{4}{S} *{4}{S}@{}}
\scalebox{0.7}{\begin{tabular}{p{0.6\columnwidth}p{0.6\columnwidth}} %{|l|l|l|l|l|l|l|l|l|l|}%
    \toprule
      \textbf{Natural} & \textbf{Synthetic}  \\
      \midrule

    \textcolor{darkgray}{Something came up} \textcolor{purple}{something came up a problem} \textcolor{darkgray}{\textbf{sorry something came up} and I cann't make it to the party} & \textcolor{darkgray}{Can’t believe I got stood up...} \textcolor{purple}{Mi} \textcolor{darkgray}{date} \textcolor{purple}{dijo} \textcolor{darkgray}{\textbf{"sorry, something came up"} like for real?!} \\

    \midrule
    
    \textcolor{purple}{Green eyes on} \textbf{replayyy}  & This song \textcolor{purple}{has me} in my feels, \textbf{replay} x100 \\

    \midrule
    
    \textcolor{purple}{I was thinking about} \textbf{texting}\textcolor{purple}{ my sister and a minute later I get a message from her} \#sisterlyconnection & \textcolor{purple}{When I'm going to} \textbf{text} \textcolor{purple}{and you show up} typing, call it telepathy or just \textcolor{purple}{good vibes} \\
    
   %\midrule
   % \textcolor{purple}{Ke me compre un carrito pa irme con mis} friends and party lol & Dude, \textcolor{purple}{tu} ex estaba en the party, y yo tipo AWKWARD. \\

\midrule

    \textcolor{purple}{I forgot all the} \textbf{passwords} \textcolor{purple}{of the Hospital \textbf{and I couldn't enter} anywhere \#PerksoftheFinals} & UGH, \textcolor{purple}{I forgot my } \textbf{password} again \textcolor{purple}{\textbf{and I cannot enter} my account...} FML \\

\midrule

    %\textcolor{purple}{Pronunciación: Muchas veces no nos entiendes porq agregamos E antes de la S en} School, Spanish, \textcolor{purple}{no es} eschool, espanish \textcolor{purple}{es} ssschool sspanish & \textcolor{purple}{Creo que} I finally got the hang of \textcolor{purple}{esto de} code-switching, it's kinda fun!\\

    \textcolor{darkgray}{Deslike} \textcolor{olive}{adicha ella punnara makkalkum nanni} & \textcolor{darkgray}{Plot had potential, but execution} \textcolor{olive}{polilla} \color{darkgray}{, disappointed.} \\
    \bottomrule
  \end{tabular}}
  \caption{Comparisons of translated natural and synthetic sentences in Spanish-English (red) and Malayalam-English (yellow). Overlapping words or phrases are highlighted in bold. The Malayalam-English data are not translated due to its low-resource nature and the lack of available translators.} \label{tab:paralleltrans}
\end{table}

\section{Implementation Details}

%\subsection{Data Details}
%Our Malayalam-English dataset used a train-dev-test split of 3542-1000-1000.

\subsection{Data Generation Details}\label{sec:datagen}
\begin{figure*}[ht]
  \includegraphics[width=\textwidth]{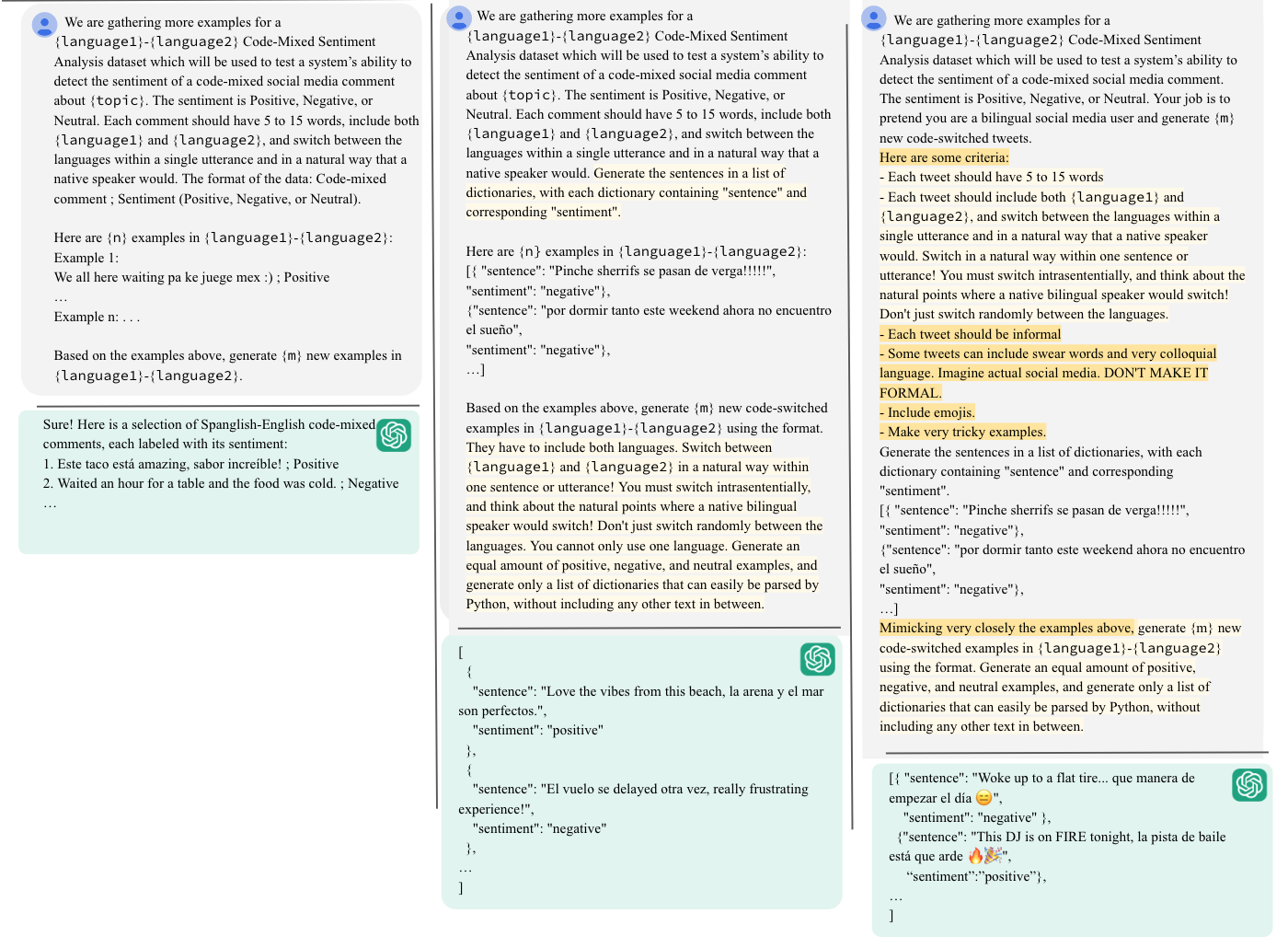}
  \caption{Prompt-tuning process, showing system input in gray, LLM sample output in teal, and iterative improvements made to our prompt highlighted in yellow. Our final prompt is shown to the right.}
  \label{fig:prompttuning}
\end{figure*}

Table \ref{fig:prompttuning} displays our prompt-tuning process, where we iteratively improved on our data generation prompt to the LLM. For all experiments, we prompted \texttt{gpt-4-1106-preview} with the OpenAI library, with temperature 0.6. For Spanish-English, we varied the number of shots \texttt{m} between \{15, 50, 150, 500\} given in our prompt with the objective to find optimal shot size. To overcome the maximum sequence length, we instructed GPT-4 to generate 50 data points and automatically repeated this process until we reached our desired dataset size. For each iteration, the prompt contained newly randomly-sampled task demonstrations from the training data. We did not post-filter the data due to its size and subjectivity. Our total synthetic data sizes were \textasciitilde53000 in Spanish-English and \textasciitilde24000 in Malayalam-English.

\subsection{Fine-tuning Details}\label{sec:finetunedetails}
Chosen based on \citet{li2022language}'s experiments, in our gradual fine-tuning approach, the synthetic data sizes were \{50000, 25000, 15000, 5000, 0\}, and each stage included 3 epochs.
For all experiments, we used the Transformers library \citep{wolf-etal-2020-transformers} to fine-tune XLM-T with a task-specific classification layer using AdamW \citep{loshchilov2019decoupled} optimizer. According to the hyperparameters of the dataset benchmark \citep{patwa-etal-2020-semeval, aguilar-etal-2020-lince} and our empirical experiments involving hyperparameter grid search, we set the highest sequence length at 40 tokens, batch size 32, weight decay 0.01, learning rate $5e^{-5}$, and epsilon $1e^{-8}$. For gradual fine-tuning, the learning rates used were \{$1e^{-6}, 2e^{-6}, 2e^{-6}, 4e^{-6}, 2e^{-6}$\}, determined through preliminary experimentation and standard grid search. We also tuned additional hyperparameters including synthetic data size, shot size, and temperature based on a standard grid search. Experiments ran on a 16GB T4 GPU. 

\section{Impact of Gradual Fine-tuning} \label{sec:Gradual}

Table \ref{tab:grad} compares F1 score for one stage of training to five stages of training using gradual fine-tuning for Spanish-English and Malayalam-English. Results marginally increase for Spanish-English while decreasing for Malayalam-English. This may be due to less suitable hyperparameters used in five stage training in comparison to one stage.

\begin{table}
\scalebox{1}{
\centering
%\begin{tabular}{@{}l *{4}{S} *{4}{S}@{}}
\begin{tabular}{lll} %{|l|l|l|l|l|l|l|l|l|l|}%
    \toprule

 \textbf{Language} & \textbf{Training} & \textbf{F1 Score}\\

    \midrule
    
    \multirow{2}{*}{Spanish-English} 
    
    & 1-Stage & 0.595\\

    & 5-Stage & 0.603\\

    \hline

    \multirow{2}{*}{Malayalam-English} 
    
    & 1-Stage & 0.843\\

    & 5-Stage & 0.718\\
    
    %hline
    
    %hline

  \end{tabular}
  }
  \caption{Comparison of F1 scores when XLM-T is fine-tuned with one stage and with five stages for each language.} \label{tab:grad}

\end{table}

\section{Instructions for Human Evaluation} \label{sec:human}

Two native Spanish-English bilingual students, who did not have knowledge of the rest of the experimentation, were each given the same 100 code-mixing texts and corresponding labels, 50 of which were randomly sampled from the natural training data and 50 of which were randomly sampled from the synthetic data. They did not know which were natural or synthetic, as the sentences were scrambled in random order.  One native Malayalam-English bilingual speaker was given 400 code-mixing texts and corresponding labels, 200 of which were randomly sampled from natural training data and 200 of which were randomly sampled from synthetic data. 

Our first Spanish-English annotator was a balanced bilingual with C2 proficiency in both English (native language) and Spanish (second language). Our second Spanish-English annotator was a balanced bilingual with C2 proficiency in Spanish (native language) and C1 proficiency in English (second language). The Malayalam-English annotator was a balanced bilingual with C2 proficiency in both Malayalam (native language) and English (second language). All annotators reported to use both languages frequently in their daily lives. The initial instructions given were:

\begin{quote}
You have been provided with a spreadsheet containing social media comments that are intended to be code-mixed in Spanish and English, though some may not be. Each comment is labeled with a sentiment—'Positive,' 'Negative,' or 'Neutral.' Your task is to evaluate each comment based on the following criteria:
Read the Sentence: Carefully review each comment.
Fill Out Ratings:
Code-mixing Naturalness: Evaluate how naturally the comment switches between Malayalam and English.
Label Accuracy: Assess whether the sentiment label ('Positive,' 'Negative,' or 'Neutral') accurately reflects the comment's connotation. If you disagree with the label, you must provide an alternative in the 'If you answered "Disagree", what would you label it?' column.
Human or Machine: Determine whether the comment was written by a human or generated artificially by a machine. 
Additional Comments (Optional): If you have further observations or concerns, please record them in the 'Additional comments' field.
Keep in mind:
Code-mixing refers to the blending of two or more languages in speech.
These comments are sourced from social media, so they may be informal, include emojis, or contain spelling errors.
If you are uncertain about your evaluation, choose the most likely option and note your concerns in the comments.
Please ensure that your evaluations are accurate and consistent across the dataset.

%Please go through this spreadsheet, and for each row, 1) Read the sentence and label, and 2) Fill out the three ratings in blue ("Code-mixing Naturalness," "Label Accuracy," \& "Human or Machine-Generated") and if applicable, the optional ones in gray (if you answered "Disagree" and "Additional comments").
\end{quote}

For "Code-mixing Naturalness," they were given the description:

\begin{quote}
Evaluate naturalness on the changing between Spanish and English. Choose between the options: "This sounds natural, like something people would actually type/say," "This sounds a bit strange/could be improved," and "This sounds unnatural/needs to be rewritten." Do not consider naturalness/strangeness of the topics discussed. Do not consider grammar/spelling mistakes unless they are extreme. Do not consider the label.
\end{quote}

It is important to note that only the first option for CM naturalness is counted as "natural" while the "strange" and "unnatural" classifications are grouped into an omnibus "unnatural" category. 
For "Label Accuracy," they were given the description:

\begin{quote}
Would you agree with the label associated with each sentence? Is a sentence labeled "positive" actually giving positive connotations? Answer with "Agree" or "Disagree." 
\end{quote}

For "Human or Machine-Generated," they were given the description:

\begin{quote}
Do you think a human wrote this or a machine wrote this? Now you can consider any and all aspects e.g. fluidity, topics, mechanics, anything.
\end{quote}

Additionally, evaluators are given the option to correct labels for which they disagreed with and to leave additional comments.

\section{Case Study on Laughter}\label{section:laugh}

\begin{table*}[htbp]
\centering
%\begin{tabular}{@{}l *{4}{S} *{4}{S}@{}}
\begin{tabular}{lp{0.5\textwidth}ll} %{|l|l|l|l|l|l|l|l|l|l|}%
    \toprule
    \textbf{Index} &
 \textbf{Sentences}
& \textbf{True Label}
& \textbf{Predicted Label}\\
\midrule
1 & jajajaj okay okay ill wait and give them to you on valentines day so it can be your cheat day
&positive
&positive\\
2 & I can imagine jajaja
&positive
&positive\\
3 & most likely jajajaj
&positive
&positive\\
4 & Girrrl I wish I had your self-esteem jaja
&neutral
&positive\\
5 & Jajajajajajajajajajajajaja ok ok ok
&neutral
&positive\\
6 & jovanigram's video JAJAJAJAJAJ
&neutral
&positive\\
7 & tb to your birthday :') jajajaja
&neutral
&positive\\
8 & Whattt Frowning Face with Open Mouth \#foreverriendome jajajjajjajaj
&neutral
&negative\\

    \bottomrule
  \end{tabular} 
  \caption{Examples of natural sentences including laughter in the test data, with true labels and predicted labels.} \label{tab:case}
\end{table*}

We investigated the use of "jajaja," shown in Table \ref{tab:case}, the Spanish version of typing laughter, which occurred frequently in both natural and synthetic data and can have positive, neutral, or negative connotations.

This case study demonstrates the challenges of CM sentiment analysis in that 1) human labels are sometimes ambiguous, 2) sentences are short, 3) the model predictions may be biased toward the positive label, and 4) emojis and symbols play an important role. Examples of ambiguity are in sentences 1 and 2, which could also be considered neutral, since sentences 5 and 6 are neutral. Sentences 2, 3, and 5 also contain very little information as compared to sentence 1, which the model had correct and shows understanding despite sentence 1's complexity. We also observe almost all positive predictions to the class imbalance as described in Section 5, where it is the model's mistake and there is fairly little ambiguity like sentence 4. For sentence 6, the model may not realize ":')" refers to a crying happy face and errs. On the other hand,for sentence 8, "Frowning Face with Open Mouth" is the English description of the emoji from the original tweet, which likely led the model to respond with negative. The change from emoji to description may also be a factor in performance worth future exploration.

\section{Generated Sentences about Code-Mixing}\label{section:aware}

Table \ref{tab:codemix} presents an intriguing observation: when asked to generate code-mixed sentences, many of the sentences ended up being about code-mixing or code-switching. In the CM sentences it was asked to generate, no theme was specified, yet out of 12865 sentences, 9 mention "code-switching," 40 mention "bilingual," 162 mention "Spanish-English," and 5 mention "French," and all discuss being skilled or having fun at code-switching. Perhaps LLM has developed somewhat of a personality, or perhaps this is due to the input instructions.

\begin{table*}
\centering
%\begin{tabular}{@{}l *{4}{S} *{4}{S}@{}}
\begin{tabular}{p{0.5\columnwidth}p{0.5\columnwidth}l} %{|l|l|l|l|l|l|l|l|l|l|}%
    \toprule
    \textbf{Sentences} & \textbf{English Translation} &
 \textbf{Labels}\\
\midrule
    \textcolor{purple}{Creo que} I finally got the hang of \textcolor{purple}{esto de} code-switching, it's kinda fun! & I think I finally got the hang of code-switching, it's kinda fun! & positive \\
    
    ¿Does this count \textcolor{purple}{como un} code-switched tweet? Asking for a friend  & Does this count as a code-switched tweet? Asking for a friend &	neutral \\

\textcolor{purple}{Ya no sé} if I should \textcolor{purple}{hablar español o inglés}, my brain is too code-switchy today & I still don't know if I should speak Spanish or English, my brain is too code-switchty today & neutral \\

Random \textcolor{purple}{pero} I started learning French y \textcolor{purple}{ahora mezclo} three languages, send help &	Random but I started learning French and already mix three languages, send help & neutral \\
    
    \bottomrule
  \end{tabular} 
  \caption{Examples of synthetic sentences mentioning CM explicitly, their translations, and their labels. Red text is in Spanish.} \label{tab:codemix}
\end{table*}

\section{Cost Analysis of Data Collection}\label{section:cost}

\subsection{Natural Data}
To estimate the cost incurred by \citet{patwa-etal-2020-semeval} of annotating 18,789 tweets using Amazon Mechanical Turk (MTurk), we first determine the number of HITs (Human Intelligence Tasks) required. Each HIT includes 10 tweets, but only 8 are for annotation purposes, with 2 serving as quality control. Thus, to annotate 18,789 tweets, we need approximately 2,349 HITs. To hire workers fluent in Spanish, HITs are required to be priced at at least \$1.00 per HIT.\footnote{\url{https://requester.mturk.com/pricing}} The total cost would then be computed as follows: 2,349 HITs multiplied by \$1.00 per HIT results in a total cost of \$2,349 USD. This estimate assumes that each HIT is completed by a single annotator and does not account for additional costs related to rejected assignments or quality control beyond the base HIT price.

Estimating the additional costs related to rejected assignments, if 30\% of all assignments were rejected and reassigned, the total cost would increase to \$3,054 USD.

These calculations use the case of \citet{patwa-etal-2020-semeval}, but it is important to consider that other works generally require more than one annotator to label each data point. Then, the previously calculated costs would double or triple depending on the number of annotators. Furthermore, \citet{patwa-etal-2020-semeval} do not release their exact price per HIT or the number of reassigned assignments, so there is high variability. Increased prices per HIT could increase costs significantly.

\subsection{Synthetic Data}
For generating synthetic data, we made requests to GPT-4 to generate 50 data points at a time. The purpose was to overcome the model's maximum sequence length. In the future, cost can be further reduced due to increasing maximum sequence length in LLMs.

To estimate the cost of generating 50,000 synthetic samples using GPT-4, we first determine the total number of tokens per request. Each request includes a prompt of 330 tokens and 15 data examples, each averaging 20.8 tokens, totaling 642 tokens for the prompt and examples. GPT-4 then generates 50 samples, each averaging 21 tokens, resulting in 1,050 tokens for the generated samples. Therefore, each request utilizes a total of 1,692 tokens. To generate 50,000 samples, we need to make 1,000 requests, resulting in a total of 1,692,000 tokens. Given GPT-4 pricing, which is \$10.00 per 1 million input tokens and \$30.00 per 1 million output tokens,\footnote{\url{https://openai.com/api/pricing/}} we can calculate the costs as follows: For the 642,000 input tokens, the cost is \$6.42, while for the 1,050,000 output tokens, the cost is \$31.50. Thus, the total cost for generating 50,000 samples is approximately \$37.92.

\end{document}